# Fuzzy Knowledge Representation, Learning and Optimization with Bayesian Analysis in Fuzzy Semantic Networks

**Momamed Nazih Omri**

*Département de Mathématiques et d'Informatique,
Institut Préparatoire aux études d'Ingénieur de Monastir,
Route de Kairouan, 5019 Monastir.
Tel. 216 3 500 273, Fax. 216 3 500 512
Nazih.Omri@ipeim.rnu.tn*

**Abstract**

This paper presents a method of optimization, based on both Bayesian Analysis technical and Gallois Lattice, of a Fuzzy Semantic Networks. The technical System we use learn by interpreting an unknown word using the links created between this new word and known words. The main link is provided by the context of the query. When novice's query is confused with an unknown verb (goal) applied to a known noun denoting either an object in the ideal user's Network or an object in the user's Network, the system infer that this new verb corresponds to one of the known goal. With the learning of new words in natural language as the interpretation, which was produced in agreement with the user, the system improves its representation scheme at each experiment with a new user and, in addition, takes advantage of previous discussions with users. The semantic Net of user objects thus obtained by these kinds of learning is not always optimal because some relationships between couple of user objects can be generalized and others suppressed according to values of forces that characterize them. Indeed, to simplify the obtained Net, we propose to proceed to an inductive Bayesian analysis, on the Net obtained from Gallois lattice. The objective of this analysis can be seen as an operation of filtering of the obtained descriptive graph.

## 1. Introduction

In order to respond to a query, an executive assistant might know very precisely the goal that the user has in mind, which means an object in a given state (the properties of the object being transformed). Moreover, even when goals are fairly well defined, it is often necessary to think about superordinate goals.

The Gallois Lattice and the fuzzy sets methods have been used to develop the "on-line instructions" mechanisms of an Intelligent Assistance System. It can be seen as a supervisor of task execution that has the "ideal user's knowledge" of (i) prerequisites of procedures, (ii) subGoals structure, and (iii) the semantic network of the elements of the device where applied procedures are used as properties, as well as (iv) the knowledge of perceptible and imperceptible effects of user's actions. With an interactive dialogue with a user, the Assistance System tries to match items provided by users in natural language with the knowledge included in the ideal user's semantic network.

The example of the technical system we consider here is Word Processor software, with Objects such as "chain-of-characters", and procedures such as "cut" or "copy". For a novice user of the software, the list of standard denominations is not obvious and he often would like to ask an expert operator about how to execute an action such as "how to rub letters" [Omri & Tijus, 99a], [Omri & Chenaina, 99b].

## 2. The Ideal Expert's and Novice User's Fuzzy Semantic Net

Construction of the Ideal Expert Knowledge starts if given a set of Tasks that are executed using elements of one technical device through procedures. The first step is the task decomposition as a hierarchy of Goal decomposition into subGoals from the level of the Goal of the task to primitive actions. The second step consists in (i) drawing up a list of possible Goals and the procedures to reach these Goals (ii) constructing the Ideal Expert Net as a classical semantic network. But, instead of using structural properties of system's interface Objects, Goals reachable with those Objects are used as properties.

However given the polysemic aspects of natural language (verbs and nouns which express goals and device objects), with the necessity of a man-machine interface that involve queries of users, the problem that is under investigation is how to match the content of a query (the label of an Object and the label of a Goal applied to this Object, as expressed by a novice user) to their corresponding items (class of Objects and Goals as properties) in the Ideal Expert Net. By answering queries of the users while they try to perform a given goal, the Expert Assistant delivers not only planning information, but also a goal structure and the knowledge of what justifies the procedure by providing the knowledge that is included in the Ideal Expert Net.

If the Assistance System does not understand the meaning of an instruction, it discusses with the user until it is able to interpret the query in its own language. With the learning of new words in natural language as the interpretation produced in agreement with the user, the system improves its representation scheme at each experiment with a new user and, in addition, takes advantage of previous discussions with users: (i) the standard Objects and recognized by the software are described in a semantic network where goals stand for properties of Objects, (ii) as the queries of an user are expressed in natural language and as they correspond more or less to these standard denominations, the system establishes fuzzy connections between its primary knowledge and the new labels of Objects or procedures expressed by the user.

## 3. How Technical System improves its representation schema

The system's capacity to interpret an unknown word using the links created between this new word and known words defines the notion of learning in our case. The main link is provided by the context of the query. When novice's query is confused with an unknown verb (goal) applied to a known noun denoting either an object in the ideal user's Network or an object in the user's Network, the system infer that this new verb corresponds to one of the known goal. With the learning of new words in natural language as the interpretation, which was produced in agreement with the user, the system improves its representation scheme at each experiment with a new user and, in addition, takes advantage of previous discussions with users. Knowledge database presents three levels or kinds of learning: we distinguish between states goals dynamics, states objects dynamics and states relationships dynamics.

## 4. Optimization of the Fuzzy Semantic Nets by Bayesian Analysis

The approach that we present in this paper is established from Procope's formalism (Poitrenaud, Richard and Tijus, 1990), based on the Gallois lattice method (Génoche & Van Mechelen, 1993) and the Bayesian formalism. The underlying idea is to end to a hierarchical structure of object users allowing to have a process of categorization by discrimination and generalization. To end to a hierarchical structure of user objects in the form of a symbolic data table, the method of the Gallois lattice is the means that we have adopted to construct the semantic user object system. This construction consists; from a symbolic table of linguistic data, to construct, in a first time the binary table (crossed system's objets with user objects are obtained by 0 and 1). In a second time the different implications between each couple of user objects. To illustrate this method, we propose to construct the semantic user objects Net corresponding to the following symbolic table (table 2).

This table allows us to construct the user objects Net with all possible implications between each couple of objects. The semantic Net of users objects obtained is not always optimal because some relationships between couple of user objects can be generalized and others suppressed according to values of forces that characterize them. Indeed, to simplify the obtained Net, we propose to proceed to an inductive Bayesian analysis, on Net obtained from Gallois lattice.

|      | Novice User 1  | Novice User 2 | Novice User 3 | Novice User 4  | Novice User 5 |
|------|----------------|---------------|---------------|----------------|---------------|
| Char | The number     | The Sign      | The letters   | The numbers    | The number    |
| Word | The numbers    | The letters   | Substantive   | The Sign       | The Sign      |
| Key  | The Characters | Substantive   | Substantive   | The Characters | The letters   |

**Table 1.** Example of symbolic table.

|      | The number | The Sign | The letters | The numbers | The Characters | Substantive |
|------|------------|----------|-------------|-------------|----------------|-------------|
| Char | 1          | 1        | 1           | 1           | 0              | 0           |
| Word | 0          | 1        | 1           | 1           | 0              | 1           |
| Key  | 0          | 0        | 1           | 0           | 1              | 1           |

**Table 2.** Gallois lattice corresponding to the table 2.

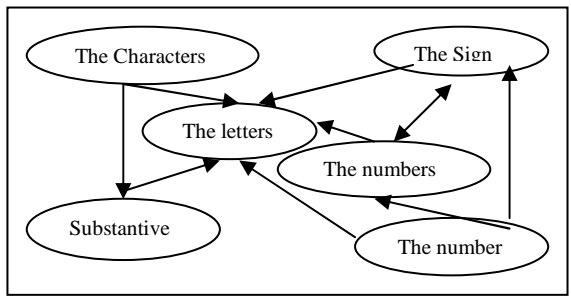

**Figure 1**. The user's objets Net corresponding on the table 2.

The principal objective of this analysis is to put to day the possible oriented dependence existing between different user objects: the knowledge of some will determine - it such or such others. To reply to this objective, we have considered the following user objects: *The number, The Sign, The letters, The numbers, The Characters* and *Substantive*. These user objects represent synonymies by novice users to designate the following system's objects: *Char, Word* and *Key*. To determine the different binary relationships between each couple, the analysis consists in study the implicative structure to each couple, then to present the totality of the implicative structures corresponding to the form of implicative graph (figure 1).

### 4.1 Descriptive inductive analysis

From observations realized on each couple of user objects, we have built the following table that presents sorting crossed in effective for each pair of user objects. Each places in table 3 represents 768 users of the software that we have put in place. For instance, in the first places, corresponding to the couple of objects *'the Sign'* and *'the number'*, 100 users have used the word *'the Sign'* to each time that they have used the word *'the number'* to designate a system's object. 30 other users have used the word *'the number'* without used the word *'the Sign'*. 85 have used the word *'the Sign'* without using the word *'the number'* and 538 remainder of the effective total have not used neither the word *'the Sign'* nor the word *'the number'* to designate system's object.

For each of these crossed sorting, we calculate the indication H of Loevinger associated to the four possible error squares. Positive indices are represented in fat (table 4). If we consider the two values-mark $h_{tend}=0.40$ and $h_{quasi}=0.60$, we have to respect next conclusions:

$$H < H_{tend} \text{ absence of q - implication}$$
$$h_{tend} \leq H \leq H_{tend} \text{ tendency to the q - implication}$$
$$H \geq H_{tend} \text{ q – implication}$$

|  | The Sign | | The letters | | The numbers | | The Characters | | Substantive | |
|---|---|---|---|---|---|---|---|---|---|---|
| The number | 100<br>85 | 30<br>553 | 50<br>143 | 80<br>495 | 49<br>100 | 81<br>538 | 38<br>70 | 92<br>568 | 66<br>50 | 64<br>588 |
| The Sign | | | 150<br>43 | 35<br>540 | 49<br>100 | 136<br>483 | 43<br>65 | 142<br>518 | 46<br>70 | 139<br>513 |
| The letters | | | | | 49<br>100 | 144<br>475 | 78<br>30 | 115<br>545 | 26<br>90 | 167<br>485 |
| The numbers | | | | | | | 49<br>59 | 100<br>560 | 29<br>87 | 120<br>532 |
| The Characters | | | | | | | | | 38<br>78 | 70<br>582 |

**Table 3:** Table of staffs crossed to each couple of user objects.

|  | The Sign | | The letters | | The numbers | | The Characters | | Substantive | |
|---|---|---|---|---|---|---|---|---|---|---|
| The number | -2,19<br>**0,45** | **0,7**<br>-0,14 | -0,53<br>**0,11** | **0,18**<br>-0,04 | -0,94<br>**0,19** | **0,22**<br>-0,05 | -1,08<br>**0,22** | **0,18**<br>-0,04 | -2,36<br>**0,48** | **0,42**<br>-0,09 |
| The Sign | | | -2,23<br>**0,71** | **0,75**<br>-0,24 | **0,57**<br>**0,12** | **0,09**<br>-0,03 | -0,65<br>**0,21** | **0,11**<br>-0,03 | -0,65<br>**0,2** | **0,11**<br>-0,04 |
| The letters | | | | | -0,3<br>**0,1** | **0,07**<br>-0,02 | -1,87<br>**0,18** | **0,31**<br>-0,1 | 0,11<br>**-0,04** | **-0,02**<br>0,01 |
| The numbers | | | | | | | -1,34<br>**0,32** | **0,22**<br>-0,05 | -0,23<br>**0,07** | **0,05**<br>-0,01 |
| The Characters | | | | | | | | | -1,33<br>**0,22** | **0,24**<br>-0,04 |

**Table 4:** Table of Loevinger's indices to each couple of user objects.

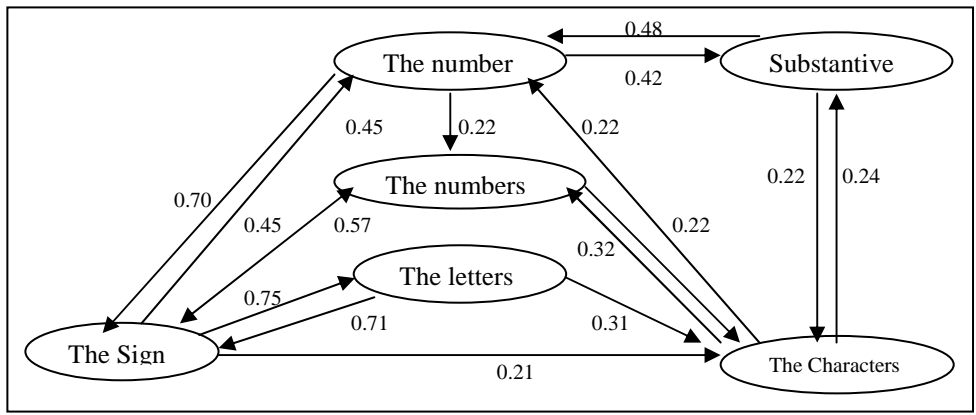

**Figure 2 :** The implicative descriptive graph of relationships with the indication H ≥ 0.20.

The suitable figure 2 emerges two possible cases. The first case, constituted following user objects *: Substantive, The number, The Sign, The letters*, and *The numbers*, positive connection following q-implication from *The number* to *The Sign, The Sign* to The letters with tendency to the equivalence, q-implication with equivalence between *The Sign* and *The numbers*, tendency to the q - implication from *The Sign* to *The numbers* and between *The number* and *Substantive* with tendency to the equivalence. The second case, constituted by the number user objects, *Substantive, The Characters* and *The Sign* presents relationships of q-exclusion and tendency to the q-exclusion.

### 4.2 Processing by the Inductive Bayesian Analysis

This stage consists in determine with the help of the AIB, observed oriented relationships descriptively that can be certified inductively, among all relationships in order that the indication H ≥ 0.20. The objective of this analysis can be seen as an operation of filtering of the obtained descriptive graph (figure 2). In order that, we are going to calculate, to each places in the table 4 above (H < 0.20), the inferior credibility limit, for a guarantee - mark δ=90, for the corresponding indication dress η. To realize these calculations, we have used a recent version of the software AIB-2 developed in the cognitive Psychology Laboratory of the University Paris VIII. Results of these calculations are presented in the following table 5. Negative values are not taken in account and therefore it does not appear in table 5.

The results of this filtering allow determining relationships that can be generalized, among the totality of observed relationships descriptively.

According to the graph of the figure 3, one can certify on the one hand, a q-implication with tendency to the equivalence between *The Sign* and *the letters* user objects and a q-implication from *The Sign* to *The number*. On the other hand, a tendency to the q-implication from *Substantive* to *The number*. For the implication from *the letters* to *The Characters* and from this last to *the numbers*, one notices that there is an absence of q-implication with tendency to the exclusion.

|                | The Sign | The letters | The numbers | The Characters | Substantive |
|----------------|----------|-------------|-------------|----------------|-------------|
| The number     | **0,634**|             | 0,168       |                | **0,36**    |
|                | **0,397**|             |             | 0,156          | **0,414**   |
| The Sign       |          | **0,698**   |             |                |             |
|                |          | **0,658**   |             | 0,135          |             |
| The letters    |          |             |             | **0,264**      |             |
| The numbers    |          |             |             | 0,171          |             |
|                |          |             |             | **0,253**      |             |
| The Characters |          |             |             |                | 0,174       |
|                |          |             |             |                | 0,159       |

**Table 5 :** Table of credibility inferior limit for each indication H with the guarantee 0.90.

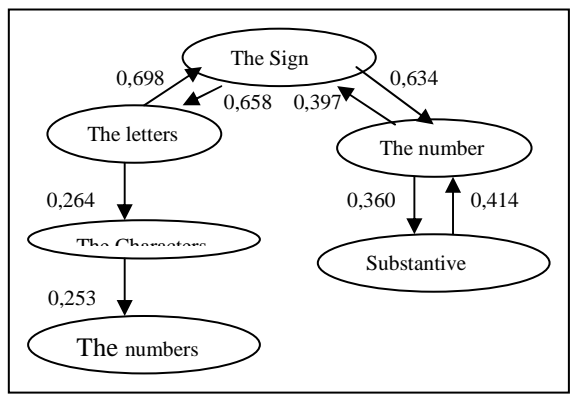

**Figure 3 :** The implicative inductive graph of relationships with the indication H ≥ 0.20.

## 5. Conclusion

Although the approach presented in this paper, that consists of an optimization of a fuzzy semantic Networks after learning, does not represent a methodology to diagnosis the goal query's novice users, allows identifying the unknown novice user request of the share of the device used. This can serve as basis for our research so as to elaborate a general methodology to diagnosis the purpose goal of the subject, applicable to a large diversity of devices. The objective being to find the totality of compatible purposes with actions of the users, the trip of such graphs facilitates grandly the research. The development of this method would have to allow a best approximation of the category of the purpose aimed by the user, and best approaches the diagnosis.

**References:**

Bordat, J.P. (1986) *Calcul pratique du treillis de Gallois d'une correspondance*. Mathématiques et Sciences humaines, 96, 31-47.

Bouchon-Meunier, B., Tijus C.A., Omri, M.N. (1993). *Fuzzy sets system for user's assistance: How Sifade diagnoses user's goal*. The Second Wold Congress on Expert Systems: "Moving towards Expert Systems Globally in the 21st Century".

Macmillian ed. CD Rom. Estoril, Lisbonne.

Card S.K., Moran T.P., Newell A. (1983). *The Psychology of Human-Computer Interaction*. Hillsdale, N.J.: Erlbaum

Génoche, A. & Van Mechelen, I. (1993). *Gallois approach to the induction of concepts*. In I. Van Mechelen, J. Hampton, R.S. Michalski, P. Theuns (Eds), *Categories and concepts: Theoretical and inductive data analysis*, (pp.287-308). London: Academic Press.

Green T.R.G., Schiele F, Payne S.J. (1992). *Formalisable models of user knowledge in human-computer interaction in working with computers*: Theory versus Outcome Van Der Veer G.C., Green T.R.G., Hoc J.M, Murray D.M. (eds) London Academic Press.

Norris, E.M. (1978) *An algorithm for computing the maximal rectangles in a binary relation*. Revue Roumaine de Mathématiques Pures et Appliquées, 23, (2), 243-250.

Omri, N.M., Tijus, C.A.(1999a). *Uncertain and approximative Knowledge Representation in Fuzzy Semantic Networks*. The Twelfth International Conference On Industrial & Engineering Applications of Artificial Intelligence & Expert Systems IEA/AIE-99. Cairo, Egypt, May 31- June 3, 1999.

Omri, N.M., Chenaina T.(1999b). *Uncertain and Approximative Knowledge Representation to Reasoning on Classification with a Fuzzy Networks Based System. The 8th IEEE International Conference on Fuzzy Systems. FUZZ-IEEE '99. Séoul. Korea.*

Omri, N.M., Chenaina T.(1999c). *Fuzzy Knowledge Representation, Learning and optimization with Bayesian Analysis Fuzzy Semantic. The 8th IEEE International Conference on Fuzzy Systems. 6th International Conference Of Neural Information Processing. ICONIP'99. Perth. Australia.*

Omri, N.M., Tijus, C.A., Poitrenaud, S. & Bouchon-Meunier, B.(1995). *Fuzzy Sets and Semantic Nets For On-Line Assistance*. Proceedings of The Eleven IEEE Conference on Artificial Intelligence Applications. Los Angeles, February 20-22.

Padgham L.(1988). *A Model and Representation for Type Information and Its Use in Reasoning with Defaults*. Proceedings of AAAI, vol. 2, p 409-414.

Poitrenaud S. (94) *The PROCOPE semantic network: an alternative to action grammars*. International Journal Of Man-Machine Studies.

Poitrenaud S., Richard J.F., Tijus C.A. (1990). *An Object-oriented semantic description of procedures for evaluation of interfaces*. 5th European Conference on Cognitive Ergonomics ECCE. Urbino, septembre 1990.

Richard J.F., Poitrenaud S. Tijus C. (1993). *Problem-solving restructuration: elimination of implicit constraints*. Cognitive Science, vol. IV, 497-529.

Storms, G., Van Mechelen, I. & De Boeck, P. (1994). Structural analysis of the intension and extension of semantic concepts. *European Journal of Cognitive Psychology, 6,* (1), 43-75.

Tauber, M. (1988). *On mental models and the user interface. Human-computer interaction in Working with Computers*: Theory versus Outcome Van Der Veer G.C., Green, T.R.G., Hoc J.M., Murray, D.M. (eds) London. Academic Press.

Tijus C.A., Poitrenaud S. (1992). *Semantic Networks of Action. NATO conference on Psychological and Educational Foundations of Technology-Based Learning Environments*. Kolymbari (Crete).

Wille, R. (1982) *Restructuring lattice theory: an approach based on hierarchies of concepts*. In O. Rival (ed), Ordered Sets. Boston: Reidel, pp:445-470.